\documentclass[10pt,twocolumn,letterpaper]{article}

\usepackage[pagenumbers]{cvpr} % To force page numbers, e.g. for an arXiv version

\usepackage{graphicx}
\usepackage{amsmath}
\usepackage{amssymb}
\usepackage{booktabs}

\usepackage{url}
\usepackage[labelformat=simple]{subcaption}
\newcommand{\tabscaleA}{0.8}

\newcommand{\R}{\mathbb{R}}

\def\modify#1{#1}
\def\reviewers#1#2#3{}
\def\modifybegin{}
\def\modifyend{}

\def\sreviewers#1#2#3{}

\usepackage[pagebackref,breaklinks,colorlinks]{hyperref}

\usepackage[capitalize]{cleveref}
\crefname{section}{Sec.}{Secs.}
\Crefname{section}{Section}{Sections}
\Crefname{table}{Table}{Tables}
\crefname{table}{Tab.}{Tabs.}

\def\cvprPaperID{*****} % *** Enter the CVPR Paper ID here
\def\confName{CVPR}
\def\confYear{2022}

\begin{document}

%%%%%%%%% TITLE - PLEASE UPDATE
\title{Model-agnostic Multi-Domain Learning\\ with Domain-Specific Adapters for Action Recognition}

\author{
    Kazuki Omi, Jun Kimata, Toru Tamaki\\
    Nagoya Institute of Technology\\
    Nagoya, Japan
}
\maketitle

%%%%%%%%% ABSTRACT
\begin{abstract}
    In this paper, we propose a multi-domain learning model for action recognition.
    The proposed method inserts domain-specific adapters between layers of domain-independent layers of a backbone network.
    Unlike a multi-head network that switches classification heads only,
    our model switches not only the heads, but also the adapters for facilitating to learn feature representations
    universal to multiple domains.
    Unlike prior works, the proposed method is model-agnostic and doesn't assume model structures unlike prior works.
    Experimental results on three popular action recognition datasets (HMDB51, UCF101, and Kinetics-400) demonstrate that
    the proposed method is more effective than a multi-head architecture
    and more efficient than separately training models for each domain.
\end{abstract}

%−−−−−−−−−−−−−−−−−−−−−−−−−−−−−−−−−−
\section{Introduction}

Video recognition tasks~\cite{Hutchinson_IEEE2021},
especially recognition of human actions, has become important in various real-world applications,
and therefore many methods have been proposed.
In order to train deep models, it is necessary to collect a variety of videos of human actions in various situations,
therefore many datasets have been proposed
\cite{kay_arXiv2017,soomro_arXiv2012,kuehne_ICCV2011}.
The distribution of a training dataset is called \emph{domain},
and the difference in distribution between two domains is called \emph{domain shift}
\cite{wilson_arXiv2020,redko_arXiv2020,wang_Elsevier2018}.
A domain is greatly characterized by the process of collecting the dataset of the domain,
therefore, it is necessary to collect training samples in several different domains for recognizing actions in various situations.
Usually recognition models are trained on a single given dataset (or domain) for performance evaluation,
but they often face to the difficulty of performing well in a cross-dataset situation,
which means that they perform well on samples of the same domain but don't well generalize on samples of other domains.

A possible approach might be
domain adaptation (DA)
\cite{wilson_arXiv2020,redko_arXiv2020,wang_Elsevier2018}.
DA approaches adapt a model trained on samples of a source domain to samples of a target domain
in order to cope with situations where training and test domains are different.
However, when there are more than two domains,
it would be better to use Multi-Domain Learning (MDL)~\cite{Rebuffi_CVPR2018,Li_CVPR2019},
which built a single model that can be used in multiple domains.
\modifybegin
Recently, many-domain problems have been attracted their attention for images
(such as
Visual Decathlon~\cite{Rebuffi_NIPS2017} and
Medical Segmentation Decathlon~\cite{Medical_Segmentation_Decathlon})
as well as videos
(Video Pentathlon~\cite{Video_Pentathlon}).
In these cases, MDL has advantages over pair-wise domain adaptation approaches
as the number of domains increases.
\modifyend

\begin{figure}[t]
  \centering

  \begin{minipage}[b]{0.6\linewidth}
    \centering
    \includegraphics[width=\linewidth]{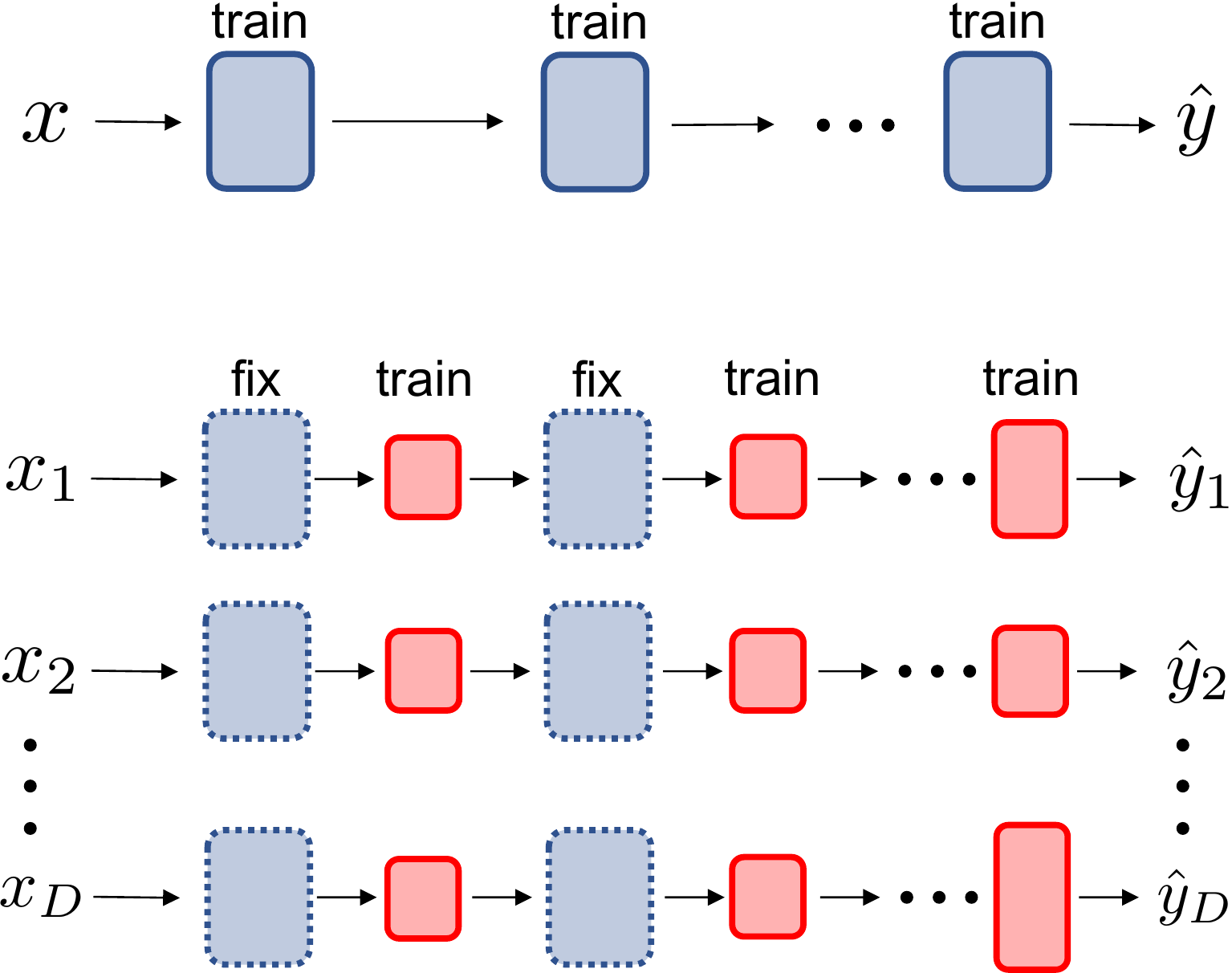}
    \subcaption{}
    \label{fig:adapter_type}
  \end{minipage}

  \medskip

  \begin{minipage}[b]{0.57\linewidth}
    \centering
    \includegraphics[width=\linewidth]{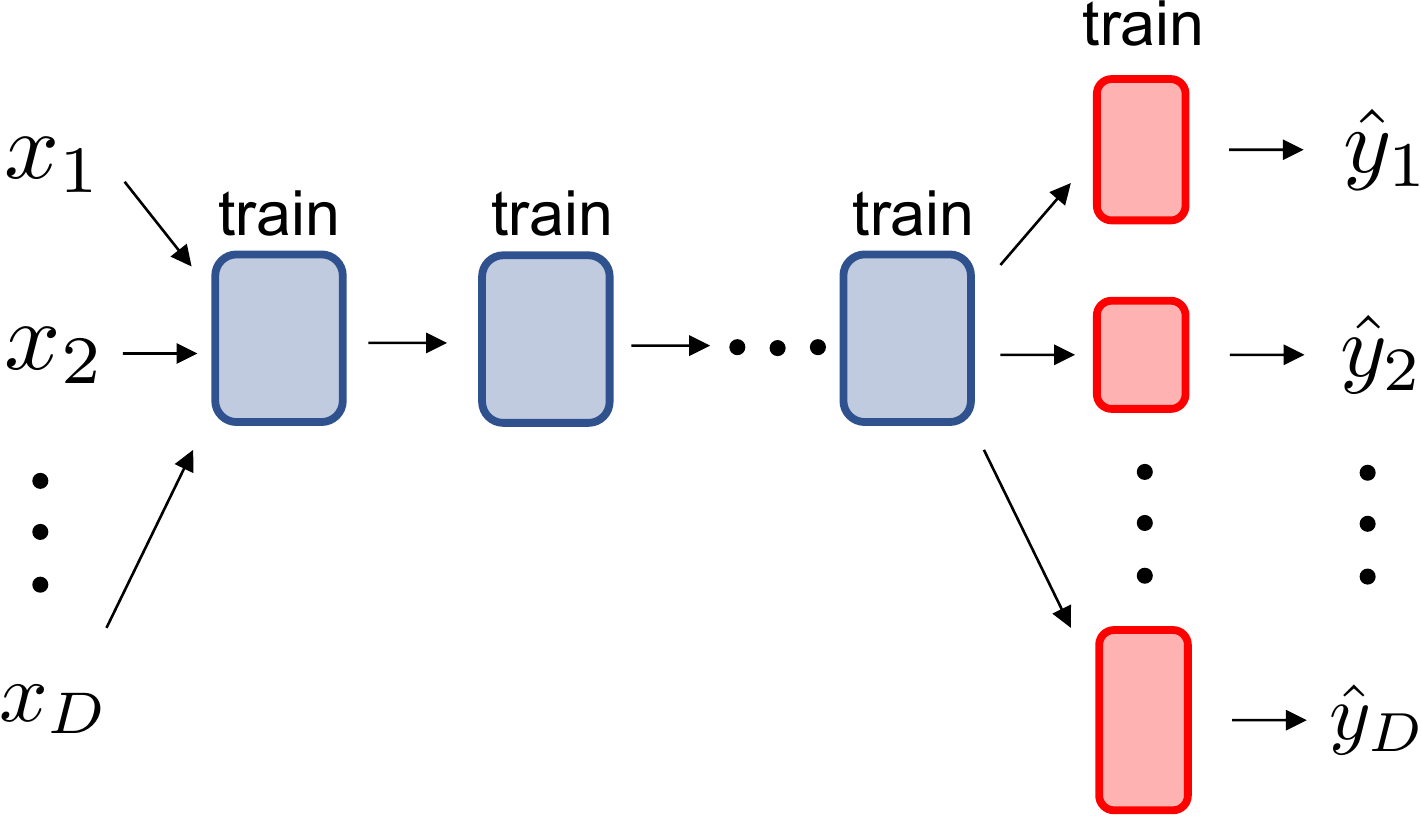}
    \subcaption{}
    \label{fig:multi_head_type}
  \end{minipage}

  \caption{Two types of multi-domain learning architectures.
    (a) Adapter type:
    after pre-training of domain-independent parameters (blue), they are fixed and domain-specific parameters (red) are trained for each domain separately.
    (b) Multi-head type:
    domain-independent (blue) and domain-specific parameters (red) are trained simultaneously for all domains.
    Note that $(x_d, \hat{y}_d)$ are input and prediction of the sample from domain $d$.
  }
  \label{fig:MDL}

\end{figure}

MDL models have two types of trainable parameters;
one is domain-\emph{independent} parameters that are shared by all domains,
and the other is domain-\emph{specific} parameters such that different domains have different ones.
A model with fewer domain-specific parameters will be computationally less expensive even when more domains are added,
while more domain-independent parameters are expected to improve the ability to represent features common for different domains.
There are two main architectures of MDL as shown in Figure~\ref{fig:MDL};
domain-specific and independent parameters are trained
separately~\cite{Rebuffi_NIPS2017,Li_CVPR2019}, or
simultaneously~\cite{Masaki_ITSC2021}.
In the former, domain-independent parameters are fixed after pre-training
and domain-specific parameters are trained on each domain separately.
In the latter, all parameters are trained on multiple domains at once.
% MDL models trained on multiple domains 
% are expected to have better performance than non-MDL models trained on each domain separately because it is expected to obtain a better feature representation.

Action recognition involves a variety of domains, however, the development of MDL models has received less attention than image recognition tasks so far, although some DA methods for action recognition have been proposed~\cite{chen_ICCV2019,pan_AAAI2020,choi_ECCV2020,Munro_CVPR2020}.
It is important to develop MDL models for video recognition tasks
because the computation cost of action recognition models often become large, and a single MDL model would be more efficient than using different models for different domains.
In this paper, we propose a new MDL model for action recognition.
The proposed method, inspired by the prior work~\cite{Li_CVPR2019},
inserts adapters with domain-specific parameters between domain-independent layers.
The contributions of this work are as follows;
\begin{itemize}

  \item We propose a method of multi-domain learning for action recognition.
        To the best of the authors' knowledge, this is the first attempt at MDL for action recognition.
        % Unlike conventional methods~\cite{Li_CVPR2019}, the source domain is not fixed and multiple domains are trained simultaneously end-to-end.

  \item Our proposed method uses adapters between layers, which can be applicable to many of existing action recognition models, unlike prior works~\cite{Rebuffi_NIPS2017,Rebuffi_CVPR2018} that restrict the model to be a ResNet with resblocks.

  \item The proposed adapter has (2+1)D convolutions that processes temporal and spatial information jointly while reducing parameters.

  \item We show experimental results with three different datasets (HMDB51, UCF101, and Kinetics400) demonstrating the effectiveness of the proposed method.

\end{itemize}

\section{Related Work}

\subsection{Action recognition and domain adaptation}

Action recognition has been an actively studied topic~\cite{Hutchinson_IEEE2021} over the last two decades, and
various models have been devised to capture the temporal information,
such as X3D~\cite{Feichtenhofer_CVPR2020} with 3D CNN,
as well as recent models~\cite{Salman_arXiv2021} based on Vision Transformer~\cite{Rohit_CVPR2019}.
However, they all require one model per domain and usually each dataset is used to train and validate models separately for performance evaluation.

Domain adaptation (DA) for action recognition has been studied to capture the difference of the appearance information as well as the temporal dynamics, which makes recognizing videos difficult compared to images.
For example,
TA$^3$N~\cite{chen_ICCV2019} introduces a domain discriminator to achieves an effective domain alignment with adversarial training.
TCoN~\cite{pan_AAAI2020} uses a cross-domain attention module to avoid frames with low information content and focus on frames commonly important both in the source and target domains.
SAVA~\cite{choi_ECCV2020} is a model that responds to human actions rather than the background for adapting domains with different backgrounds.
MM-SADA~\cite{Munro_CVPR2020} performs adaptation for each of RGB and optical flow domains.
These DA approaches however don't handle more than two domains.

\subsection{Multi-domain learning}

To handle multiple domains, an approach similar to multi-task learning would be taken, that is, using multi-heads~\cite{Masaki_ITSC2021}.
As shown in Fig.\ref{fig:MDL}(b),
the model has a single feature extractor used for all domains
and multiple classification heads for each domain.
In this case, the feature extractor has domain-independent parameters,
while each head has its own domain-specific parameters.
However, as more domains are involved, it will become more difficult for a single extractor to extract universal features for multiple domains,
particularly for complex video domains.

Another approach is to insert adapters in a backbone network~\cite{Rebuffi_NIPS2017,Rebuffi_CVPR2018,Li_CVPR2019} as shown in Fig.\ref{fig:MDL}(a).
First, the backbone model is pre-trained to fix the domain-independent parameters. Then adapters, which are domain-specific parameters, are inserted to the backbone network. Finally, the modified network is trained on each domain.
One drawback of this approach is that
the backbone network is assume to have a ResNet structure
to insert adapters in parallel or series inside the resblocks~\cite{Rebuffi_NIPS2017,Rebuffi_CVPR2018}.
Hence it is difficult to apply the adapter approach to other models, even though a variety of pre-trained models~\cite{Feichtenhofer_CVPR2020,feichtenhofer_ICCV2019} are currently available.
% the inserted adapters should be model-independent.
%
To alleviate this issue,
CovNorm~\cite{Li_CVPR2019} doesn't assume model structures
and inserts model-agnostic adapters between layers.
However, the training is not end-to-end because adapters need the dimensionality reduction of features offline by principal component analysis.

In contrast, our method doesn't assume the model structure, like as~\cite{Li_CVPR2019}, while the training is done in an end-to-end manner.
In addition, the proposed method fine-tunes all the parameters during the training with multiple domains, whereas
prior works using adapters~\cite{Rebuffi_NIPS2017,Rebuffi_CVPR2018}
have fixed pre-trained domain-independent parameters (of the backbone network) during the training with multiple domains.
% Therefore, performance degradation occurs in domains that differ significantly from the source domain.

\begin{figure}[t]
  \centering

  \includegraphics[width=\linewidth]{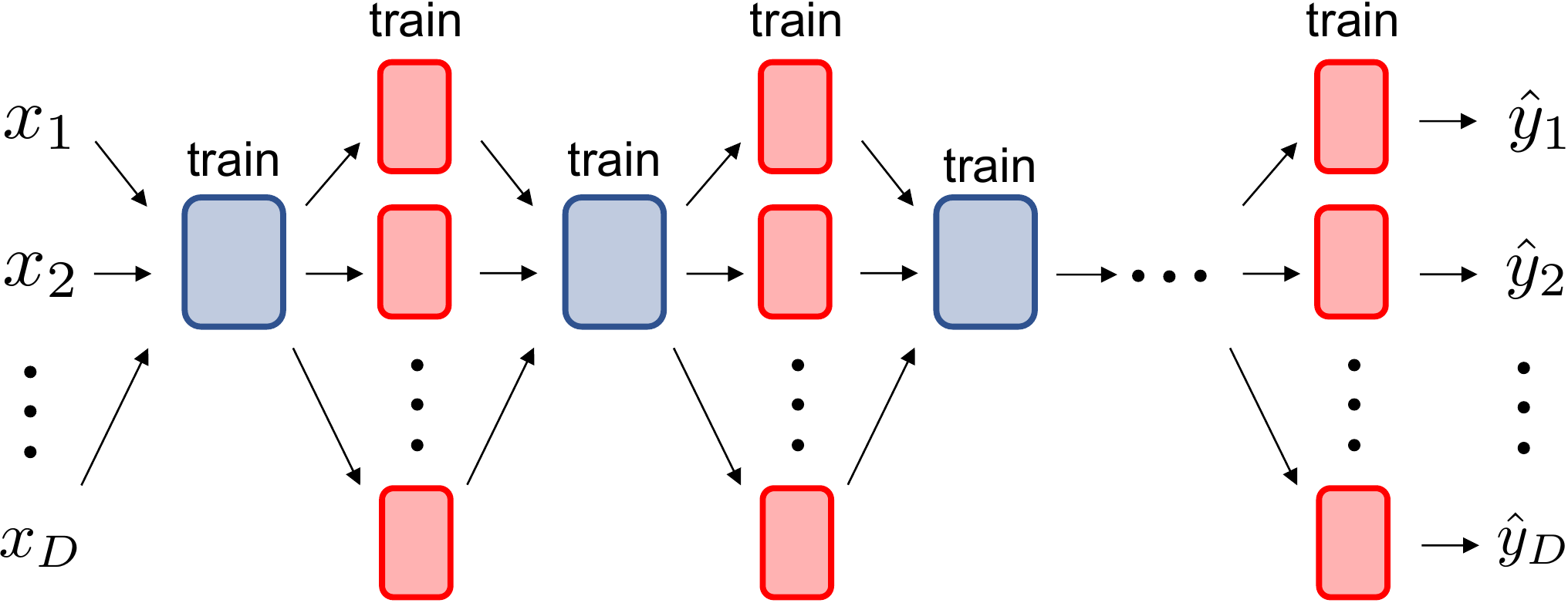}

  \caption{An overview of our proposed method,
    which trains domain-independent backbone layers (blue) and domain-specific adapters (red) for all domains simultaneously.
    Unlike other multi-head and adapter models, this model has adapters for each domain between backbone layers.
  }
  \label{fig:method}

\end{figure}

\begin{figure*}[t]
  \centering

  \hfil
  \begin{minipage}[t]{0.25\linewidth}
    \centering
    \includegraphics[keepaspectratio, scale=0.3]{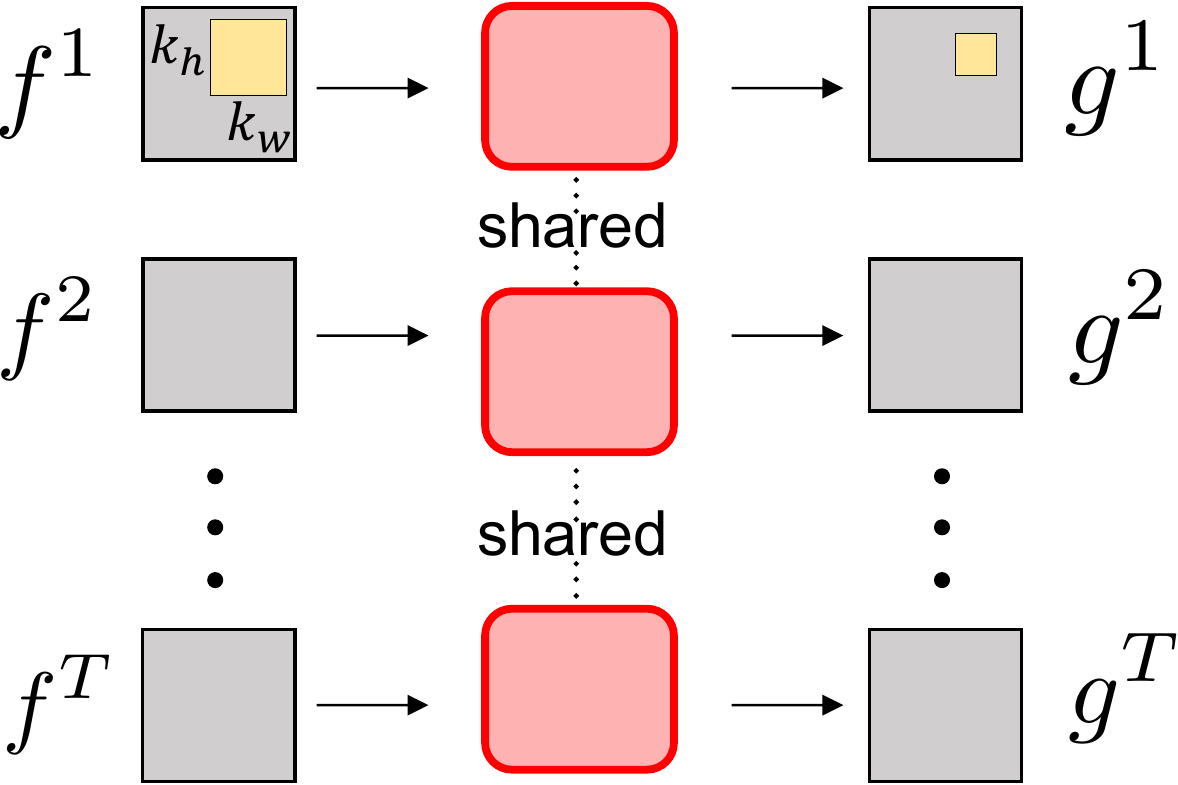}
    \subcaption{}
  \end{minipage}
  \hfil
  \begin{minipage}[t]{0.3\linewidth}
    \centering
    \includegraphics[keepaspectratio, scale=0.25]{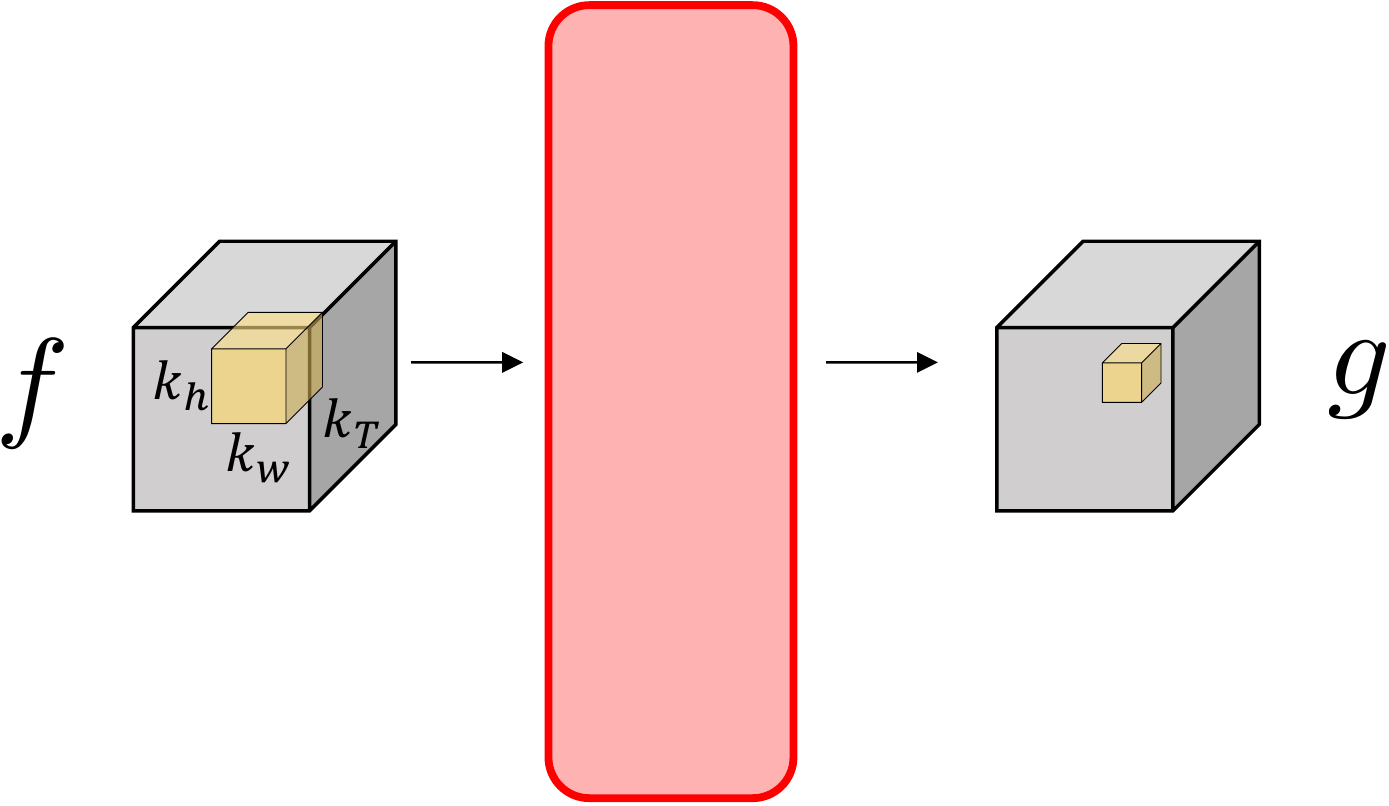}
    \subcaption{}
  \end{minipage}
  \hfil
  \begin{minipage}[t]{0.35\linewidth}
    \centering
    \includegraphics[keepaspectratio, scale=0.3]{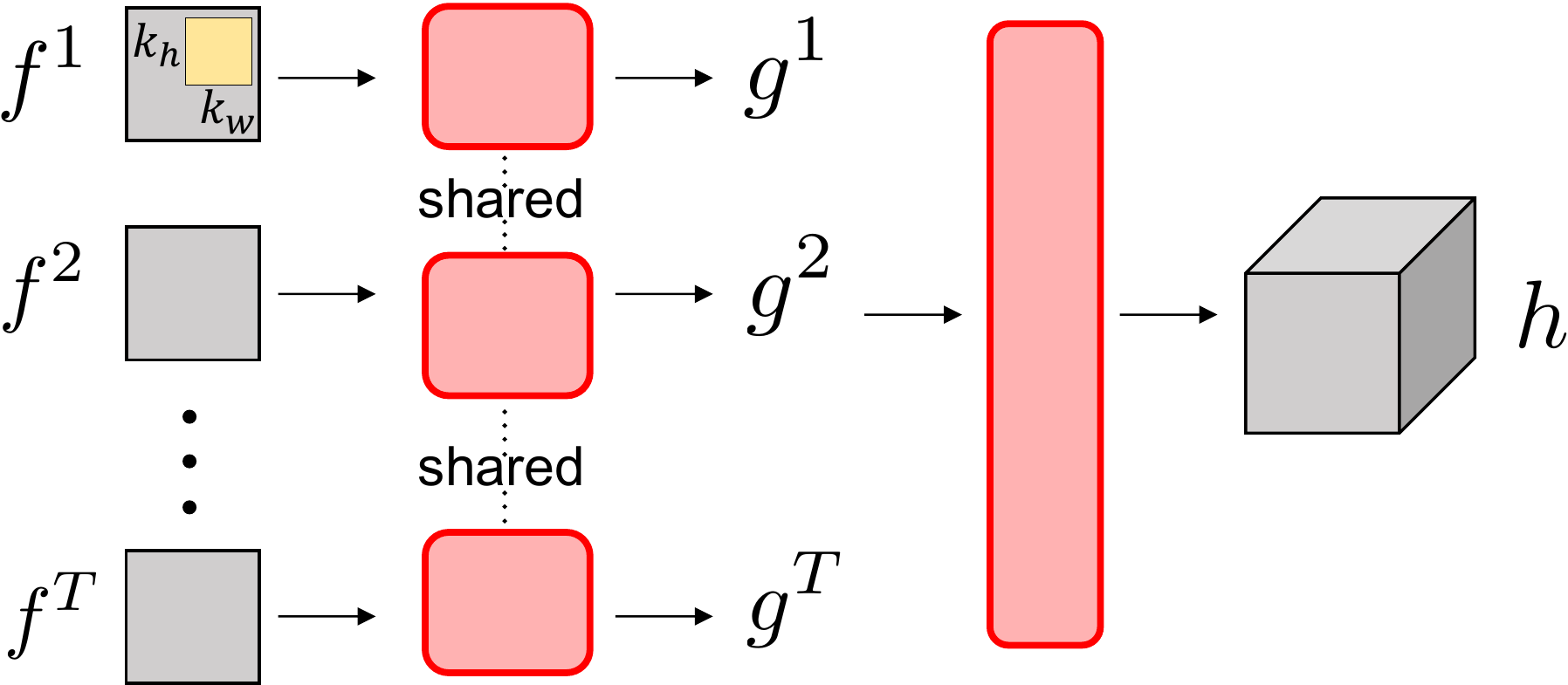}
    \subcaption{}
  \end{minipage}
  \hfill

  \caption{Three types of adapters;
    (a) frame-wise 2D convolutions,
    (b) 3D convolution, and
    (c) (2+1)D convolution.
  }
  \label{fig:adapter}

\end{figure*}

\section{Method}

\subsection{Architecture}

Figure~\ref{fig:method} shows the overview of the proposed method.
The core idea is the use of adapters between layers like as CovNorm~\cite{Li_CVPR2019},
but different adapters are used for different domains
like as classification heads in a multi-head network~\cite{Masaki_ITSC2021}.
First, we pre-train a backbone model that has $N$ layers (or stages, blocks),
each of which is shown as blue modules in Fig.\ref{fig:method}.
This is the same with the top of Fig.\ref{fig:MDL}(a) where only the backbone model is shown.

Let $M^\ell$ be the $\ell$-th layer in the backbone,
that takes input $f^\ell \in \R^{T \times C^\ell \times H^\ell \times W^\ell}$
and output $f^{\ell+1} \in \R^{T \times C^{\ell+1} \times H^{\ell+1} \times W^{\ell+1}}$.
Here $H^\ell$ and $W^\ell$ are spatial dimensions (width and height) of $f^\ell$ with $C^\ell$ channels.
The first layer takes an input video clip
$x=f^1 \in \R^{T \times 3 \times H^1 \times W^1}$,
where $T$ is the number of frames in the clip,
assuming that the layers doesn't change the temporal extent of the input.
The last layer $M^L$ predicts the softmax score $\hat{y} \in [0,1]^N$
of $N$ categories.
Using these notations, the backbone network is assume to be a type of stacking layers;
\begin{align}
  \hat{y} = M^L (M^{L-1} ( \cdots M^2 ( M^1 (x) ) \cdots ) ).
  \label{eq:module_stacking_model}
\end{align}
Note that this type is widely used in many architectures,
such as 3D ResNet~\cite{hara_ICCV2017} and X3D~\cite{Feichtenhofer_CVPR2020}.

Next, we insert adapter $A_d^\ell$ between
layers $M^\ell$ and $M^{\ell+1}$ for $\ell = 1, \ldots, L - 2$.
Thus the adapter takes $f^{\ell+1}$ and output a transformed feature $g^{\ell+1}$ of the same shape,
which is then passed to the next layer $M^{\ell+1}$.
Here $d$ is the index of domains $d \in \{1,\dots,D\} = \mathcal{D}$.
This means that we use different adapters $A_d^\ell$ for different domain $d$;
\begin{align}
  \hat{y}_d = M_d^L ( M^{L-1} (A_d^{L-2} \cdots M^2 (A_d^1 ( M^1 (x_d) ) ) \cdots ) ) ).
\end{align}
Note that we don't insert adapters just before the head $M_d^L$ because the head itself is domain-specific.

As shown in Fig.\ref{fig:method}, when the network input is a sample $x_d = f_d^1$ from domain $d$,
then the data passes through domain-specific adapters $A_d^1, A_d^2, \ldots, A_d^{L-2}$
between the domain-independent backbone layers $M^1, M^2, \ldots, M^{L-1}$ during the forward and backward computations.
At the end of the network,
there are multiple heads $M_d^L$, each for domain $d$,
predicting scores $\hat{y}_d \in [0, 1]^{N_d}$ where $N_d$ is the number of categories in domain $d$.
This is the same as the multi-head architecture (Fig.\ref{fig:MDL}(b)), but our method switches not only the heads but also the adapters for each domain
depending on from which domain the sample comes.

\subsection{Loss}

Then, we train the whole network, that is, all of the domain-specific parameters
(adapters $A_d^\ell$ and classification heads $M_d^L$)
as well as the domain-independent parameters (backbone layers $M_\ell$).
Let $(x_{i,d}, y_{i,d})$ is $i$-th input-output pair of domain $d$.
\modify{Note that domain $d$ of each sample is given.}
Then, we minimize the following cross entropy loss;
\begin{align}
  L
   & =
  E_{d \sim \mathcal{D}} E_{ x, y \ \sim d} [ L_\mathit{CE,d}( \hat{y}, y ) ]
  \\
   & \approx
  \sum_d \sum_i  L_\mathit{CE,d}( \hat{y}_{i,d}, y_{i,d} ),
  \label{eq:empirical_loss}
\end{align}
assuming that the domain is sampled from a discrete uniform distribution.

Naively implementing this empirical loss is however inefficient when different samples come from different domains,
causing the network to switch adapters for each sample.
Instead, it would be more efficient if all samples in a batch come from the same domain
because the forward computation of the batch uses adapters of the same domain without adapter switching.
Therefore, we introduce the following loss to minimize for a sampled batch $\{x_i, y_i\}_{i=1}^B$;
\begin{align}
  L
   & =
  E_{d \sim \mathcal{D}} E_{ \{x_i, y_i\}_{i=1}^B \sim d}
  \left[ \sum_{i=1}^B L_\mathit{CE,d}( \hat{y}_i, y_i ) \right],
\end{align}
where $B$ is the batch size.

In our implementation, a domain is selected sequentially (rather than randomly),
and a batch is sampled from the domain,
then the loss of the domain is computed.
The gradient is updated only after batches sampled from all domains have been used for backward computations.
In other words, parameters are only updated once after every $D$ backward computations.

\subsection{Spatio-temporal adapters}
\label{sec:adapters}

We proposed to use the following three types of adapters
(i.e., 2D, 3D, and (2+1)D) that spatially and temporally transform features.

\noindent\textbf{Frame-wise 2D conv.}
The 2D adapter performs convolutions for each frame separately.
Let $f \in R^{T \times C \times H \times W}$ be the input feature,
and $f^t \in R^{C \times H \times W}$ be the feature of $t$-th frame for $t = 1, \ldots, T$.
2D adapters perform 2D convolution $A_\mathrm{2D}$ to each frame separately;
\begin{align}
  g^t = A_\mathrm{2D} \otimes f^t,
\end{align}
where $\otimes$ represent convolutions.
This is implemented by 3D convolutions $A_\mathrm{3D}$
with the kernel of size $R^{C \times 1 \times k_h \times k_w}$;
\begin{align}
  g = A_\mathrm{3D} \otimes f,
  \label{eq:3Dadapter}
\end{align}
to produce the output $g$.

\noindent\textbf{3D conv.}
Unlike the 2D adapter that doesn't transform features temporally,
the 3D adapter uses 3D convolution on the 3D video volume (Figure~\ref{fig:adapter}(b)).
An adapter $A_\mathrm{3D}$ is applied as in the same with Eq.\eqref{eq:3Dadapter}
with the kernel of size $R^{C \times k_t \times k_h \times k_w}$.

\noindent\textbf{(2+1)D conv.}
3D convolution is expected to model the temporal information of actions
because it considers both spatial and temporal dimensions simultaneously.
However, as the number of adapters increases with the number of domains,
adapters are required having fewer parameters.
Inspired by separable convolution~\cite{tran_CVPR2018,Qiu_ICCV2017,Xie_2018_ECCV},
we introduce (2+1)D convolution adapters that use two convolutions in series; one for spatial and the other for temporal.
First, frame-wise 2D convolutions with the kernel of size size $R^{C \times 1 \times k_h \times k_w}$ are applied;
\begin{align}
  g^t = A_\mathrm{2D} \otimes f^t, \ \text{for} \ t=1,\ldots,T,
\end{align}
then a 1D convolution with the kernel of size $R^{C \times k_t \times 1 \times 1}$ aggregates the outputs of $T$ frames
along the temporal direction;
\begin{align}
  g = A_\mathrm{1D} \otimes [g^1, g^2, \ldots, g^T].
\end{align}

\begin{figure}[t]
  \centering

  \includegraphics[width=0.6\linewidth]{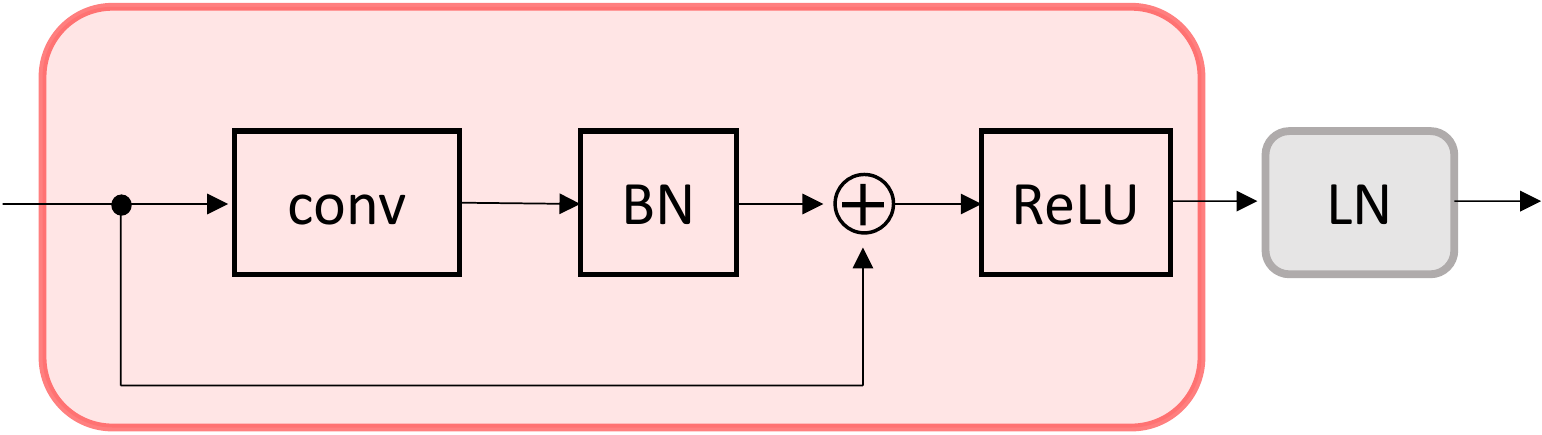}

  \caption{The structure of adapters.
    The ``conv'' layer is either 2D, 3D, or (2+1)D convolutions.
  }
  \label{fig:adp_structure}

\end{figure}

\subsection{Adapter structure}

Figure~\ref{fig:adp_structure} shows the structure of adapters.
Each adapter has a batch normalization (BN) after either of 2D, 3D, or (2+1)D convolutions described above,
followed by skip connection and ReLU.
In Fig.\ref{fig:adp_structure}, the red plate represents an adapter $A_d^\ell$, that is switched for each domain $d$.
In addition, we place a layer normalization (LN) as additional domain-independent parameters
after the output of these adapters.
Adapters output domain-specific features, which may differ for each domain.
We expect LN to make the domain-specific adapter outputs more domain-independent
for facilitating the training of the next layer.

\section{Experiments}

We show experimental results using three domains,
and compare the proposed method with multi-head and non-MDL approaches.

\subsection{Setting}

\paragraph{Datasets}

HMDB51~\cite{kuehne_ICCV2011} consists of 3.6k videos in the training set and 1.5k videos in the validation set, with 51 human action categories.
Each video is collected from movies, web, Youtube, etc., and the shortest video is less than 1 second and the longest is about 35 seconds, while most videos are between 1 and 5 seconds long, with an average length of 3.15 seconds.
% There are three splits for 3570 training and 1530 validation videos, and
The first split was used in this experiment.

UCF101~\cite{soomro_arXiv2012} consists of 9.5k videos in the training set and 3.5k videos in the validation set, with 101 human action categories.
Each video was collected from Youtube, and the video length is 1 second for the shortest and 30 seconds for the longest,
while most videos are between 3 and 10 seconds in length, with an average length of 7.21 seconds.
There are three splits for training and validation, and we report the performance of the first split as it is usually used.
% The number of videos for training and validation are (9537, 3783) for the first split.

Kinetics400~\cite{kay_arXiv2017} consists of 22k videos in the training set, 18k videos in the validation set,
and 35k videos in the test set, with 400 human action categories.
Each video was collected from Youtube and trimmed to a 10 second long segment corresponding to one of the action categories.

\paragraph{Model}

\begin{figure}[t]

  \centering

  \includegraphics[width=\linewidth]{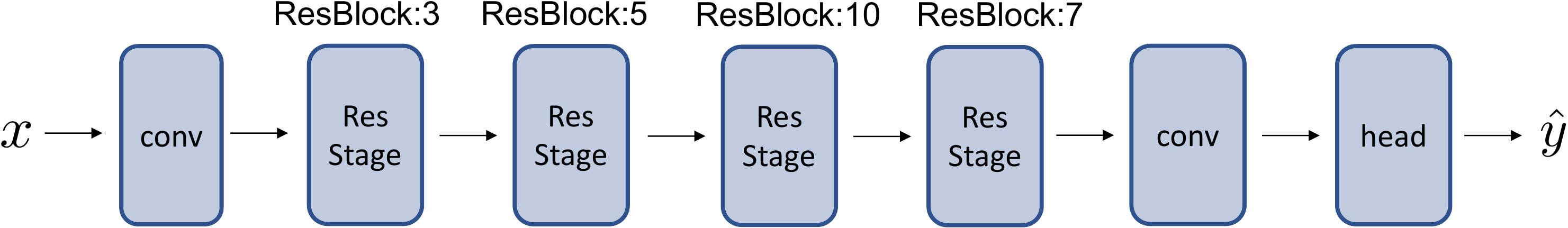}\\
  (a)\\
  \vspace{0.5cm}
  \includegraphics[width=\linewidth]{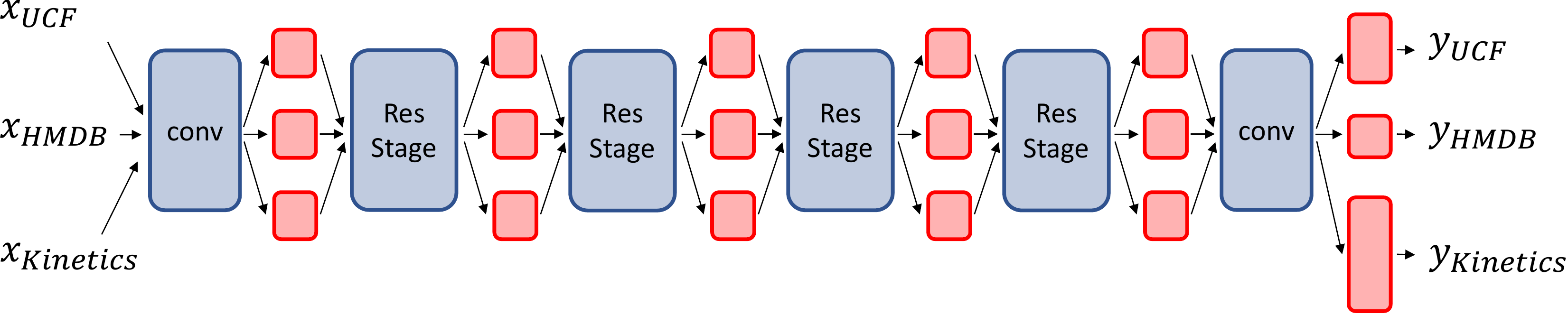}\\
  (b)

  \caption{Structures of
    (a) the backbone X3D-M, and
    (b) our model with adapters and heads for each domain.
  }
  \label{fig:model}

\end{figure}

We used X3D-M~\cite{Feichtenhofer_CVPR2020} pre-trained on Kinetics400 as the backbone network.
Figure~\ref{fig:model}(a) shows the structure of X3D-M,
which has the stem conv block $M^1$, followed by four ResBlock stages $M^2, \ldots, M^5$ and a conv block $M^6$,
and finally a classification head $M^7$ (hence $L=7$).
The proposed model used in the experiments is shown in Fig.\ref{fig:model}(b).
This model has $D=3$ classification heads $M_d^7$ at the end of the network,
and five adapters per domain $A_d^\ell$ (for $\ell=1,\ldots,5$) between the backbone modules.

We used the following adapter parameters.
For frame-wise 2D conv, the kernel size was
% $k_h \times k_w = 1 \times 1$,
\modify{$k_h \times k_w = 3 \times 3$}.
% which is similar to existing adapter models~\cite{Rebuffi_NIPS2017,Rebuffi_CVPR2018,Li_CVPR2019}.
For 3D conv, the kernel size was $k_t \times k_h \times k_w = 3 \times 3\times 3$.
For (2+1)D conv, the kernel size for spatial convolution was $k_h \times k_w = 3 \times 3$ and for temporal convolution $k_t = 3$.

\modifybegin
In the following experiments,
unless otherwise denoted,
we used the (2+1)D adapters,
which were inserted to all five locations
(the ``all'' row in Tab.\ref{Tab:adp_place}),
for the proposed method (Fig.\ref{fig:method},
the ``train\&train'' row in Tab.\ref{Tab:fix_shared_param})
with X3D-M as a backbone.
\modifyend

\paragraph{Training}

Training video files of different datasets differ in fps. Therefore, we used the following protocol, following \texttt{pytorchvideo}~\cite{fan2021pytorchvideo}.
From one video in the training set, we randomly extracted consecutive frames corresponding to a specified duration
starting from a randomly decided position,
and created a clip by sampling 16 frames uniformly from the extracted frames.
We used the duration of about 2.67 seconds (corresponding to 80 frames in 30 fps) because of the setting of X3D-M (using 16 frames with a stride of 5 frames).
The short sides of extracted frames were randomly resized to  [224, 320] pixels and resized while maintaining the aspect ratio.
Then we randomly cropped $224 \times 224$ pixels and flipped them horizontally with a probability of 50\%.

The backbone X3D-M model were pre-trained on Kinetics400, so were the domain-independent parameters.
We trained from scratch the adapters and heads (domain-specific parameters), as well as LN layers (added as domain-independent parameters).

The term ``epoch'' doesn't make sense because we train the models on three datasets simultaneously and different datasets have different number of samples.
Therefore, in the experiments,
we trained models for 42,000 iterations, corresponding to 14,000 iterations for each dataset.
The batch size was set to 32, therefore the effective numbers of training epochs were about 2 for Kinetics400, 48 for UCF101, and 128  for HMDB51.
The input clips for training were taken from the three datasets in turn for each batch.
In other words, the first batch of 32 clips was taken from the first dataset,
the second batch was taken from the second dataset,
the third batch was taken from the third dataset, and so on, for 42,000 batches.
When training a batch of dataset $d$, the batch is passed through adapters $A_d^\ell$ and head $M_d^L$,
as well as domain-independent layers $M^\ell$,
to compute the loss $L_\mathit{CE,d}$.
The gradient is back-propagated using layers and adapters only used in the forward computation.
However, parameters are not updated until the gradients of batches of three datasets have been back-propagated.
In this experiment, parameters were updated once every three batches, each from three different datasets.

We used an SGD optimizer with momentum of 0.9.
The initial learning rate was set to 0.001 and reduced to 1/10 at 8,000 and 12,000 iterations.

\paragraph{Inference}

In validation, we performed a multi-view test as in prior works~\cite{Nonlocal}.
For each video in the validation set,
we repeated the clip sampling 10 times to sample 10 clips.
Then we resized the frames while maintaining the aspect ratio so that the short side was 256 pixels, and cropped to $224\times224$ at the right, center, and left.
This generated 30 clips (30 views), and we averaged these results to compute a single prediction score.

\subsection{Results}

\begin{table}[t]
  \centering

  \caption{The top-1 performance with different adapter types.
    %   for the validation sets of HMDB51, UCF101, and Kinetics400.
    %   The first three rows show results of three backbone models with no adapters trained on each dataset separately.
    \modify{Number of parameters are also shown for the backbone model (base), heads, and adapters.}
  }
  \label{Tab:acc_adp}

  \scalebox{\tabscaleA}{

    \modifybegin

    \begin{tabular}{@{}c@{\hspace{.3em}}|@{\hspace{.3em}}c@{\hspace{.5em}}c@{\hspace{.5em}}c@{\hspace{.3em}}|@{\hspace{.3em}}c@{\hspace{.3em}}|@{\hspace{.3em}}c@{\hspace{.5em}}c@{\hspace{.5em}}c@{\hspace{.5em}}c@{}}
              &       &       &       &         & \multicolumn{4}{c}{params (M)}                       \\
      adapter & HMDB  & UCF   & K400  & average & total                          & base & head & adap. \\ \hline
      2D      & 73.07 & 95.93 & 69.94 & 79.65   & 5.45                           & 2.97 & 1.13 & 1.34  \\
      (2+1)D  & 74.77 & 96.25 & 69.84 & 80.29   & 5.89                           & 2.97 & 1.13 & 1.79  \\
      3D      & 75.03 & 95.77 & 70.08 & 80.29   & 8.12                           & 2.97 & 1.13 & 4.02  \\
    \end{tabular}

    \modifyend
  }

  %   \begin{tabular}{c|ccc|c|c}
  % %   adapter & UCF  & HMDB  & K400  & average & params \\ \hline
  % %     ---    & --- & 75.62 & --- & --- & 3.08M \\ % 3,079,173 = 2,974,674 + 104,499
  % %     ---    & 97.28 & --- & --- & --- & 3.18M \\ % 3,181,623 = 2,974,674 + 206,949
  % %     ---    & --- & --- & 72.43 & --- & 3.79M \\ % 3,794,274 = 2,974,674 + 819,600
  % %     \hline
  % %     2D     & 95.43 & 73.07 & 68.70 & 79.07 & 4.26M \\ % 4,257,786
  % %     (2+1)D & 96.25 & 74.77 & 69.84 & 80.29 & 5.89M \\ % 5,893,626
  % %     3D     & 95.77 & 75.03 & 70.08 & 80.29 & 8.12M \\ % 8,121,594
  % adapter & HMDB  & UCF   & K400  & average & params \\ \hline
  % % ---     & 75.62 & ---   & ---   & ---     & 3.08M  \\
  % % ---     & ---   & 97.28 & ---   & ---     & 3.18M  \\
  % % ---     & ---   & ---   & 72.43 & ---     & 3.79M  \\ \hline
  % % 2D      & 73.07 & 95.43 & 68.70 & 79.07   & 4.26M  \\
  % \modify{2D}    & \modify{73.07} & \modify{95.93} & \modify{69.94} & \modify{79.65} & \modify{?.??M} \\ 
  % (2+1)D         & 74.77 & 96.25 & 69.84 & 80.29   & 5.89M \\
  % 3D             & 75.03 & 95.77 & 70.08 & 80.29   & 8.12M 
  % \end{tabular}

\end{table}

\subsubsection{Adapter types}

First we compare three types of adapters.
Table~\ref{Tab:acc_adp} shows the performances for each adapter type.
As expected, 3D and (2+1)D adapters performed better than 2D adapters
because of the ability to model the temporal information.
In the following experiments, we used (2+1)D conv because it has fewer parameters while both 3D and (2+1)D performed similarly.

\subsubsection{Fixing or fine-tuning domain-independent parameters}

In the prior works with adapters~\cite{Rebuffi_NIPS2017,Rebuffi_CVPR2018,Li_CVPR2019},
the domain-independent parameters of the backbone were pre-trained on some domain,
then fixed during training with multiple domains.
In contrast, our model fine-tunes those parameters to jointly train with adapters.
Table~\ref{Tab:fix_shared_param} shows the performance comparison of these two settings.
The first row shows the performance of our model with adapters inserted,
but the domain-independent backbone layers were not trained during multi-domain learning.
As expected, the performance is better when all parameters are trained jointly,
indicating that training adapters only is insufficient to support multiple domains.
The backbone layer should extracts more generic domain-independent features,
which makes the feature transformation with adapters more effective.

\begin{table}[t]
  \centering
  \caption{
    The top-1 performances by fixing or fine-tuning domain-independent layers.
    \modify{
      Note that the number of trainable parameters are shown.
      The first row corresponds to the architecture shown in
      Fig.\ref{fig:adapter_type},
      and the second row to Fig.\ref{fig:method}.}
  }
  \label{Tab:fix_shared_param}

  \scalebox{\tabscaleA}{
    \modifybegin

    \begin{tabular}{@{}c@{\hspace{.3em}}c@{\hspace{.3em}}|c@{\hspace{.3em}}c@{\hspace{.3em}}c@{\hspace{.3em}}|@{\hspace{.3em}}c@{\hspace{.3em}}|@{\hspace{.3em}}c@{\hspace{.3em}}c@{\hspace{.3em}}c@{\hspace{.3em}}c@{}}
               &                   &       &       &       &         & \multicolumn{4}{c}{params (M)}                       \\
      $M^\ell$ & $A_d^\ell, M_d^L$ & HMDB  & UCF   & K400  & average & total                          & base & head & adap. \\ \hline
      fix      & train             & 73.07 & 95.19 & 67.54 & 78.60   & 2.91                           & ---  & 1.13 & 1.79  \\
      train    & train             & 74.77 & 96.25 & 69.84 & 80.29   & 5.89                           & 2.97 & 1.13 & 1.79  \\
    \end{tabular}
    \modifyend
    %   \begin{tabular}{cc|ccc|cc}
    %     % $M^\ell$ & $A_d^\ell, M_d^L$ & UCF & HMDB  & Kinetics & average & params \\ \hline
    %     % fix   & train   & 95.19 & 73.07 & 67.54 & 78.60 & 2.91M \\ % 2,918,952
    %     % train & train   & 96.25 & 74.77 & 69.84 & 80.29 & 5.89M \\ % 5,893,626
    % $M^\ell$ & $A_d^\ell, M_d^L$ & HMDB  & UCF   & K400 & average & params \\ \hline
    % fix      & train             & 73.07 & 95.19 & 67.54    & 78.60   & 2.91M  \\
    % train    & train             & 74.77 & 96.25 & 69.84    & 80.29   & 5.89M 
    % \end{tabular}
  }

\end{table}

\subsubsection{Adapter locations in the backbone}
\label{sec:adapter_locations}

Here we investigate the different configurations of adapter insertion.
Table~\ref{Tab:adp_place} shows the performances by changing positions where we insert adapters in the backbone model.
``Early-$x$'' used adapters $A_d^1, \ldots, A_d^x$ between the early layers of the backbone,
while ``late-$x$'' inserted adapters $A_d^{\ell-2-(x-1)}, \ldots, A_d^{\ell-2}$ between the late layers.
These configurations also have domain-specific heads $M_d^L$,
but ``multi-head'' is the case using only the heads but no adapters.
``All'' is the full model that uses all of the adapters.

On average,
the multi-head type shows the least performance, indicating that
that domain-specific parameters are needed not only at the final heads, but also between layers as adapters.
The best performance was obtained by early-1, which has the first adapter $A_d^1$ only in addition to the heads as domain-specific parameters.
As the positions of adapters inserted in the backbone becomes deeper, the performance deteriorates gradually,
which is consistent with the fact that the multi-head has domain-specific parameters only at very the end of the network.

The prior work~\cite{Rebuffi_CVPR2018} has reported that
better performances were obtained when adapters were inserted in the late layers rather than the early layers.
The differences between our work and theirs are that
videos come from similar datasets,
all the parameters are trained jointly,
and a specific backbone model is not assumed.
Three datasets in this experiments have similar categories, and most videos were taken from third-person views.
Therefore adapters in the early layers might be enough to transform low-level temporal information of videos in these datasets.
We would have different results with other datasets of first-person views, such as SSv2~\cite{goyal_arXiv2017} and Epic-Kitchen~\cite{damen_ECCV2018},
which are significantly different domains.
Another factor may be the use of X3D pre-trained on Kinetics as the backbone.
Its structure was explored in a greedy way,
so adding adaptors and heads for multiple domains may be suboptimal.

\begin{table}[t]
  \centering
  \caption{The top-1 performance with different adapter configurations for the validation sets.
    \modify{The row ``multi head'' corresponds to the architecture shown in Fig.\ref{fig:multi_head_type}.}
  }
  \label{Tab:adp_place}

  \modifybegin

  \scalebox{\tabscaleA}{
    \begin{tabular}{@{}c@{\hspace{.3em}}|@{\hspace{.3em}}c@{\hspace{.5em}}c@{\hspace{.5em}}c@{\hspace{.3em}}|@{\hspace{.3em}}c@{\hspace{.3em}}|@{\hspace{.3em}}c@{\hspace{.5em}}c@{\hspace{.5em}}c@{\hspace{.5em}}c@{}}
                 &       &       &       &         & \multicolumn{4}{c}{params (M)}                       \\
      \# config  & HMDB  & UCF   & K400  & average & total                          & base & head & adap. \\ \hline
      early-1    & 74.77 & 96.38 & 71.00 & 80.72   & 4.13                           & 2.97 & 1.13 & 0.02  \\
      early-3    & 74.64 & 96.19 & 70.75 & 80.53   & 4.23                           & 2.97 & 1.13 & 0.13  \\
      late-3     & 74.90 & 95.90 & 70.45 & 80.42   & 5.85                           & 2.97 & 1.13 & 1.75  \\
      late-1     & 73.27 & 96.03 & 70.86 & 80.29   & 5.44                           & 2.97 & 1.13 & 1.33  \\
      multi-head & 73.99 & 96.25 & 70.62 & 79.98   & 4.11                           & 2.97 & 1.13 & ---   \\
      all        & 74.77 & 96.25 & 69.84 & 80.29   & 5.89                           & 2.97 & 1.13 & 1.79  \\
    \end{tabular}

    \modifyend
    %   \begin{tabular}{c|ccc|cc} 
    %     % config & UCF & HMDB  & K400 & average & params \\ \hline
    %     % early-1     & 96.38 & 74.77 & 71.00 & 80.72 & 4.11M \\ 
    %     % early-3     & 96.19 & 74.64 & 70.75 & 80.53 & 4.15M \\ 
    %     % late-3      & 95.90 & 74.90 & 70.45 & 80.42 & 4.69M \\ 
    %     % late-1      & 96.03 & 73.99 & 70.86 & 80.29 & 4.55M \\ 
    %     % multi head  & 96.25 & 73.07 & 70.62 & 79.98 & 4.11M \\ % 4,105,722 \\
    %     % all         & 96.25 & 74.77 & 69.84 & 80.29 & 5.89M \\ % (2+1)D 5,893,626 \\
    % config     & HMDB  & UCF   & K400  & average & params \\ \hline
    % early-1    & 74.77 & 96.38 & 71.00 & 80.72   & 4.13M  \\
    % early-3    & 74.64 & 96.19 & 70.75 & 80.53   & 4.23M  \\
    % late-3     & 74.90 & 95.90 & 70.45 & 80.42   & 5.85M  \\
    % late-1     & 73.99 & 96.03 & 70.86 & 80.29   & 5.44M  \\
    % multi head & 73.07 & 96.25 & 70.62 & 79.98   & 4.11M  \\
    % all        & 74.77 & 96.25 & 69.84 & 80.29   & 5.89M 
    % \end{tabular}
  }

\end{table}

\subsubsection{Number of domains}

In MDL, the number of domains is the important factor.
\modify{
  Table~\ref{tab:acc_domain_all_config} shows the results when different numbers of domains were involved
  for the ``all'' configuration in Tab.\ref{Tab:adp_place},
  and Table~\ref{tab:acc_domain_early-1_config} for the ``early-1'' configuration.
}
Rows of ``\# domains 1'' are the cases using a single domain,
which means that the network have adapters between layers and a single head,
and is trained on the domain only.
The performance of HMDB increases as more domains are used, demonstrating that MDL is beneficial for smaller datasets
by leveraging information from other larger datasets.
In contrast, performances of UCF and Kinetics decreases when other datasets were used.
In particular, performances dropped significantly when HMDB, the smallest one, was used jointly
as shown in rows of ``\# domains 2''.
This issue of dataset sizes may caused by several factors.
Currently we assume that the domain was sampled from a uniform distribution, regardless of the dataset size,
as in Eq.\eqref{eq:empirical_loss}.
Also we minimize the sum of losses of different datasets without any weights.
We would investigate the effects of these factors in future,
by introducing non-uniform domain distributions or importance sampling.

\begin{figure}[t]
  \centering

  \begin{minipage}[b]{0.48\linewidth}
    \centering
    \includegraphics[width=\linewidth]{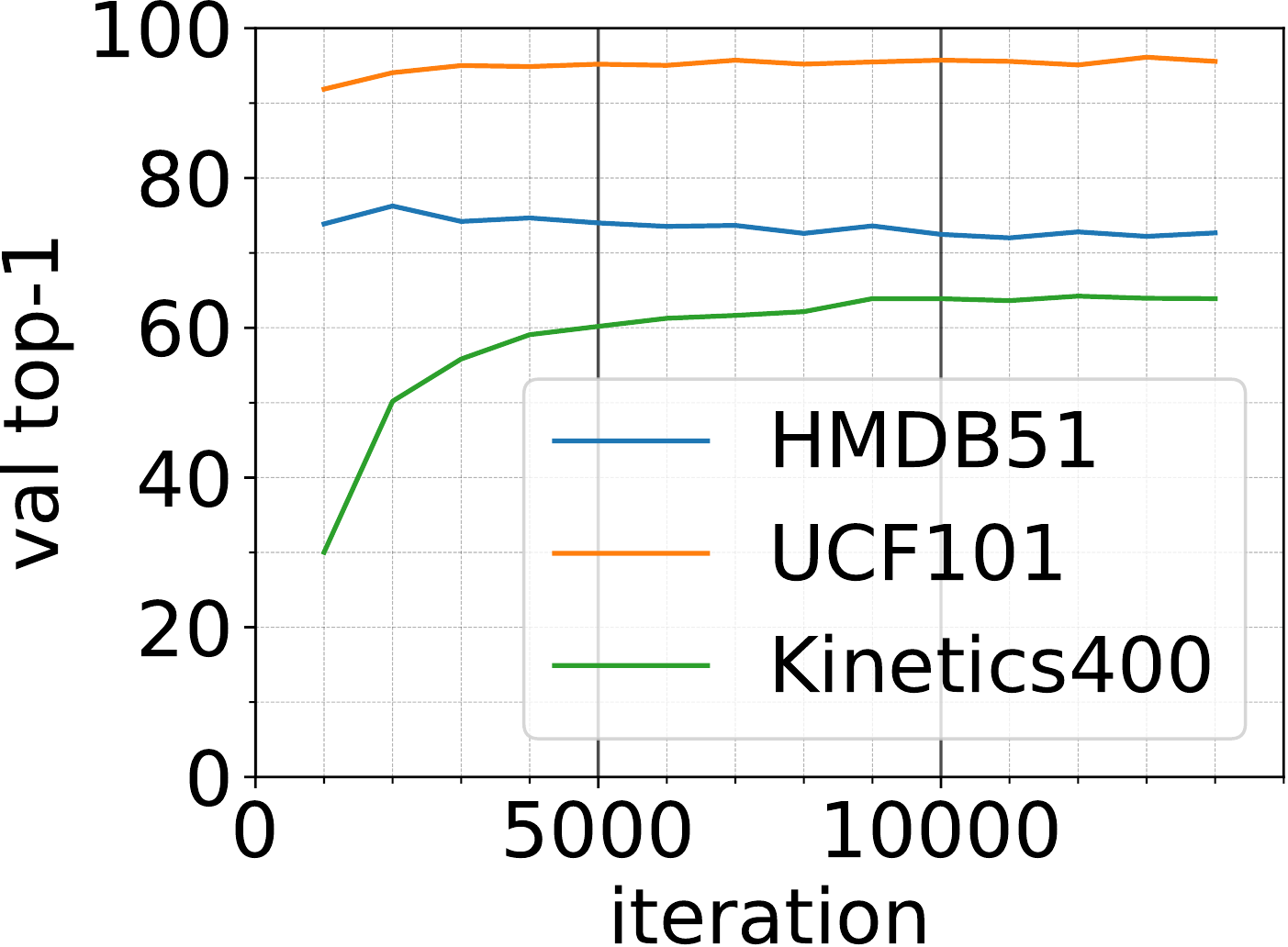}
    \subcaption{}
  \end{minipage}
  \hfill
  \begin{minipage}[b]{0.48\linewidth}
    \centering
    \includegraphics[width=\linewidth]{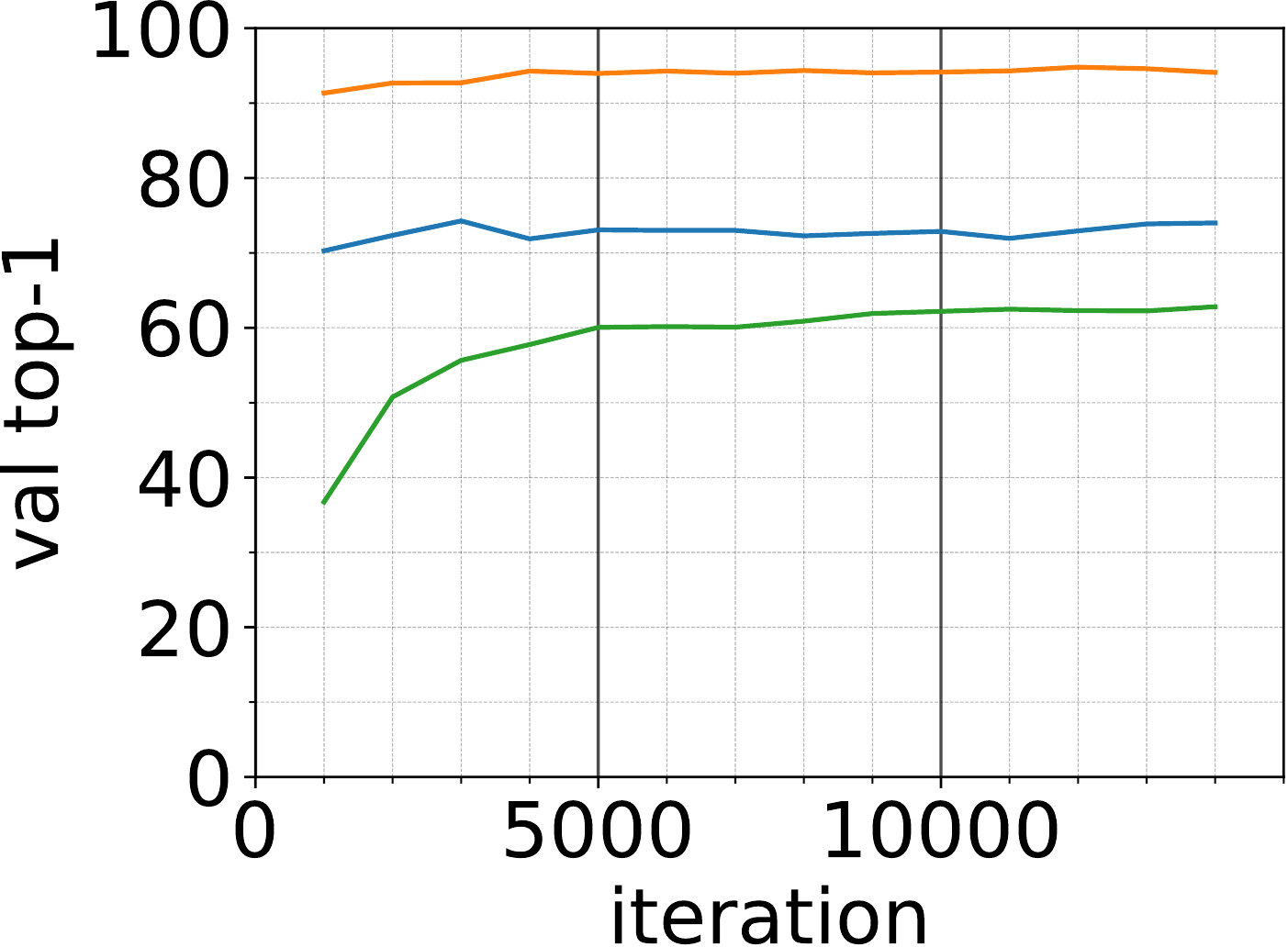}
    \subcaption{}
  \end{minipage}

  \caption{Performance over training epochs of using
    (a) a single domain, or
    (b) three domains.
  }
  \label{fig:loss_acc}

\end{figure}

Figure~\ref{fig:loss_acc} shows the performance of the validation sets of three datasets
when the network was trained on a single domain (``\# domains 1'' in Tab.\ref{tab:acc_domain_all_config})
or on three domains (``\# domains 3'').
Note that the validation performance is of a single view (not 30 views as mentioned before),
and horizontal axes Fig.\ref{fig:loss_acc}(a) and (b) should be interpreted differently
as in Fig.\ref{fig:loss_acc}(b) a single iteration refers to a single gradient update after back-propagation of three domains.
The performance of HMDB deteriorates as training progresses when trained on a single domain,
but this is not the case when trained on multiple domains.
This is in agreement with the observation in Tab.\ref{tab:acc_domain_all_config} above.

\modifybegin
Note that Tab.\ref{tab:acc_domain_all_config}
also shows the performances of the backbone network without any adapters
in rows with \mbox{``---''} in the domain column.
This shows that our model with adapters doesn't work better than the backbone itself,
even for a single domain. This might be due to the increase of the parameters to be trained
while fixing the training iterations.
But we should note that three backbone networks are needed for three domains to train separately
and have more parameters (about 10M in total), % 10,055,070
whereas our method requires a single model of fewer parameters (about 5.8M).
\modifyend

\begin{table}[t]

  \centering
  \caption{Effect of the number of domains on the top-1 performance
    \modify{with the X3D backbone of (a) the all and (b) early-1 configurations.
      Note that ``---'' in the left-most column
      shows the cases using no adapters.}
  }
  \label{Tab:acc_domain}

  \modifybegin

  \begin{minipage}{\linewidth}\centering
    \scalebox{\tabscaleA}{
      \begin{tabular}{c|c@{\hspace{.8em}}c@{\hspace{.8em}}c|c@{\hspace{.8em}}c@{\hspace{.5em}}c@{\hspace{.5em}}c@{}}
               &       &       &       & \multicolumn{4}{c}{params (M)}                       \\
        \# dom & HMDB  & UCF   & K400  & total                          & base & head & adap. \\ \hline
        ---    & 75.62 & ---   & ---   & 3.08                           & 2.97 & 0.10 & ---   \\
        ---    & ---   & 97.28 & ---   & 3.18                           & 2.97 & 0.21 & ---   \\
        ---    & ---   & ---   & 72.43 & 3.79                           & 2.97 & 0.82 & ---   \\ \hline
        1      & 73.27 & ---   & ---   & 3.68                           & 2.97 & 0.10 & 0.60  \\
        1      & ---   & 96.88 & ---   & 3.78                           & 2.97 & 0.21 & 0.60  \\
        1      & ---   & ---   & 71.80 & 4.39                           & 2.97 & 0.82 & 0.60  \\ \hline
        2      & 74.58 & 95.90 & ---   & 4.48                           & 2.97 & 0.31 & 1.19  \\
        2      & 74.25 & ---   & 70.21 & 5.10                           & 2.97 & 0.92 & 1.19  \\
        2      & ---   & 96.34 & 70.77 & 5.19                           & 2.97 & 1.03 & 1.19  \\ \hline
        3      & 74.77 & 96.25 & 69.84 & 5.89                           & 2.97 & 1.13 & 1.79
      \end{tabular}
    }
    \subcaption{}
    \label{tab:acc_domain_all_config}
  \end{minipage}

  \begin{minipage}{\linewidth}\centering
    \scalebox{\tabscaleA}{
      \begin{tabular}{c|c@{\hspace{.8em}}c@{\hspace{.8em}}c|c@{\hspace{.8em}}c@{\hspace{.5em}}c@{\hspace{.5em}}c@{}}
               &       &       &       & \multicolumn{4}{c}{params (M)}                       \\
        \# dom & HMDB  & UCF   & K400  & total                          & base & head & adap. \\ \hline
        1      & 74.18 & ---   & ---   & 3.09                           & 2.97 & 0.10 & 0.01  \\
        1      & ---   & 96.67 & ---   & 3.19                           & 2.97 & 0.21 & 0.01  \\
        1      & ---   & ---   & 72.00 & 3.80                           & 2.97 & 0.82 & 0.01  \\ \hline
        2      & 73.73 & 96.27 & ---   & 3.30                           & 2.97 & 0.31 & 0.01  \\
        2      & 74.44 & ---   & 70.99 & 3.91                           & 2.97 & 0.92 & 0.01  \\
        2      & ---   & 96.56 & 71.48 & 4.02                           & 2.97 & 1.03 & 0.01  \\ \hline
        % 3        & 74.31 & 95.93 & 70.15 &  4.13 & 2.97 & 1.13 & 0.02
        3      & 74.77 & 96.38 & 71.00 & 4.13                           & 2.97 & 1.13 & 0.02
      \end{tabular}
    }
    \subcaption{}
    \label{tab:acc_domain_early-1_config}
  \end{minipage}

  \modifyend

\end{table}

\modifybegin

\subsection{Using another backbone model}

Our proposed method is model-agnostic
and applicable to any models that have a structure like Eq.\eqref{eq:module_stacking_model}.
We have used X3D-M~\cite{Feichtenhofer_CVPR2020} in the experiments above,
and here we show results of 3D ResNet~\cite{hara_ICCV2017}
pre-trained on Kinetics400.
It has the same structure with X3D-M;
it has the stem conv block $M^1$,
four ResBlock stages $M^2, \ldots, M^5$,
a conv block $M^6$,
and a head $M^7$ (hence $L=7$).
As like in Fig.\ref{fig:model}(b),
this model has a head and five adapters per domain for the ``all'' configuration.

Results shown in Tab.\ref{Tab:acc_domain_3d_resnet_all_config} show that
the all configuration considerably increases the number of parameters
because
3D ResNet have larger channels (1024) than X3D-M (192),
which leads to 150 times more parameters of adapters for 3 domains,
and the deterioration of the performance.

In contrast, the early-1 configuration shown in Tab.\ref{Tab:acc_domain_3d_resnet_early-1_config}
have fewer parameters and better performance.
Again, this observation supports the discussion in Sec.\ref{sec:adapter_locations}
that the early layers play an important role for transforming
low-level temporal information of different domains.

% Results are shown in Tab.\ref{Tab:acc_domain_3d_resnet},
% which is similar to Tab.\ref{tab:acc_domain_all_config},
% but there are two differences;
% using three domains at the same time with a single model
% doesn't look beneficial for the smallest HMDB,
% and the number of parameters considerably increased.

% The increase of parameters is due to the difference in the number of channels in the backbone;
% for example, the last (fourth) ResBlock $M^5$ of X3D-M and 3D ResNet have
% 192 and 1024 channels respectively.
% This difference leads to 50 times more parameters of adapters in 3D ResNet than in X3D-M.
% However, since the number of iterations during training was the same,
% the training might be insufficient.
% Our future work includes to mitigate this issue by introducing a bottle neck in the adapter structure to reduce the dimension and parameters.
\modifyend

\begin{table}[t]

  \centering
  \caption{
    \modify{
      Effect of the number of domains on the top-1 performance
      with the 3D ResNet backbone of (a) the all and (b) early-1 configurations.
      Note that ``---'' in the left-most column
      shows the cases using no adapters.
    }
  }
  \label{Tab:acc_domain_3d_resnet}

  \modifybegin

  \begin{minipage}{\linewidth}\centering
    \scalebox{\tabscaleA}{
      \begin{tabular}{c|c@{\hspace{.8em}}c@{\hspace{.8em}}c|c@{\hspace{.8em}}c@{\hspace{.5em}}c@{\hspace{.5em}}c@{}}
               &       &       &       & \multicolumn{4}{c}{params (M)}                         \\
        \# dom & HMDB  & UCF   & K400  & total                          & base  & head & adap.  \\ \hline
        ---    & 73.07 & ---   & ---   & 31.74                          & 31.63 & 0.10 & ---    \\
        ---    & ---   & 96.11 & ---   & 31.84                          & 31.63 & 0.21 & ---    \\
        ---    & ---   & ---   & 70.95 & 32.45                          & 31.63 & 0.82 & ---    \\ \hline
        1      & 66.34 & ---   & ---   & 81.93                          & 31.63 & 0.10 & 50.19  \\
        1      & ---   & 94.18 & ---   & 82.03                          & 31.63 & 0.21 & 50.19  \\
        1      & ---   & ---   & 69.12 & 82.65                          & 31.63 & 0.82 & 50.19  \\ \hline
        2      & 65.56 & 92.89 & ---   & 132.33                         & 31.63 & 0.31 & 100.38 \\
        2      & 64.71 & ---   & 67.99 & 132.94                         & 31.63 & 0.92 & 100.38 \\
        2      & ---   & 93.52 & 68.27 & 133.04                         & 31.63 & 1.03 & 100.38 \\ \hline
        3      & 63.33 & 93.29 & 67.80 & 183.34                         & 31.63 & 1.13 & 150.58
      \end{tabular}
    }
    \subcaption{}
    \label{Tab:acc_domain_3d_resnet_all_config}

  \end{minipage}

  \begin{minipage}{\linewidth}\centering
    \scalebox{\tabscaleA}{
      \begin{tabular}{c|c@{\hspace{.8em}}c@{\hspace{.8em}}c|c@{\hspace{.8em}}c@{\hspace{.5em}}c@{\hspace{.5em}}c@{}}
               &       &       &       & \multicolumn{4}{c}{params (M)}                        \\
        \# dom & HMDB  & UCF   & K400  & total                          & base  & head & adap. \\ \hline
        1      & 73.59 & ---   & ---   & 31.78                          & 31.63 & 0.10 & 0.04  \\
        1      & ---   & 95.66 & ---   & 31.88                          & 31.63 & 0.21 & 0.04  \\
        1      & ---   & ---   & 70.77 & 32.49                          & 31.63 & 0.82 & 0.04  \\ \hline
        2      & 71.57 & 94.82 & ---   & 32.02                          & 31.63 & 0.31 & 0.08  \\
        2      & 71.18 & ---   & 69.98 & 32.63                          & 31.63 & 0.92 & 0.08  \\
        2      & ---   & 96.14 & 70.22 & 32.74                          & 31.63 & 1.03 & 0.08  \\ \hline
        3      & 71.90 & 96.30 & 69.39 & 32.88                          & 31.63 & 1.13 & 0.11
      \end{tabular}
    }

    \subcaption{}
    \label{Tab:acc_domain_3d_resnet_early-1_config}

  \end{minipage}

  \modifyend

  % #params of resnet, each of five adapters:
  % 37,256
  % 591,176
  % 2,361,928
  % 9,442,376
  % 37,759,048
  % total of adapters: 50,191,784

  % params of slow_r50     32,454,096 (including K400 head)
  % only stem of slow_r50  31,634,896 (no head)
  % K400 head   : 2048*400=819,200
  % UCF101 head : 2048*101=206,848
  % HMDB51 head : 2048*51 =104,448

\end{table}

\section{Conclusion}

In this paper, we propose a multi-domain learning model for action recognition that inserts domain-specific adapters between layers.
The proposed method enables an end-to-end learning with multiple domains simultaneously,
and experimental results showed that the proposed methods is more effective than a multi-head architecture,
and more efficient than training a model for each domain separately.
Our future work includes the further investigation on
the inserting locations and structures of adapters to facilitate extracting common features across different domains,
as well as domain-specific features suitable for each domain.
In addition, other datasets~\cite{goyal_arXiv2017,damen_ECCV2018} which are largely different from datasets used in the experiments of this paper,
are planned to be used for further experiments.

\section*{Acknowledgement}
\noindent
This work was supported in part by JSPS KAKENHI Grant Number JP22K12090.

    %%%%%%%%% REFERENCES
    {\small
        \bibliographystyle{ieee_fullname}
        \bibliography{mybib}
    }

\end{document}